# Intra-operative tumour margin evaluation in breast-conserving surgery with deep learning

Wei-Chung Shia[1*], Yu-Len Huang[2*§], Yi-Chun Chen[2], Hwa-Koon Wu[3], Dar-Ren Chen[4§]

[1] Molecular Medicine Laboratory, Department of Research, Changhua Christian Hospital, Changhua, Taiwan

[2] Department of Computer Science, Tunghai University, Taichung, Taiwan

[3] Department of Medical Imaging, Changhua Christian Hospital, Changhua, Taiwan

[4] Comprehensive Breast Cancer Center, Changhua Christian Hospital, Changhua, Taiwan

§Corresponding author

*These authors share the same contributions

Email addresses:

    WCS: weichung.shia@gmail.com

    YLH: ylhuang@thu.edu.tw

    YCC: hyl.thu@gmail.com

    HKW: 17597@cch.org.tw

    DRC: 115045@cch.org.tw

# Abstract

**Background**

A positive margin may result in an increased risk of local recurrences after breast retention surgery for any malignant tumour. In order to reduce the number of positive margins would offer surgeon real-time intra-operative information on the presence of positive resection margins. This study aims to design an intra-operative tumour margin evaluation scheme by using specimen mammography in breast-conserving surgery.

**Method**

Total of 30 cases were evaluated and compared with the manually determined contours by experienced physicians and pathology report. The proposed method utilizes image thresholding to extract regions of interest and then performs a deep learning model, i.e. SegNet, to segment tumour tissue. The margin width of normal tissues surrounding it is evaluated as the result.

**Results**

The desired size of margin around the tumor was set for 10 mm. The smallest average difference to manual sketched margin (6.53 mm ± 5.84). In the all case, the SegNet architecture was utilized to obtain tissue specimen boundary and tumor contour, respectively. The simulation results indicated that this technology is helpful in discriminating positive from negative margins in the intra-operative setting.

**Conclusions**

The aim of proposed scheme was a potential procedure in the intra-operative measurement system. The experimental results reveal that deep learning techniques can draw results that are consistent with pathology reports.

# Background

The breast-conserving therapy (BCT) is a recommended choice for the treatment of early stage invasive breast cancer, and only cancerous tissue plus a rim of normal tissue can be cleaned without removing the breast (Benda et al., 2004). The surgery can not only remove the tumour but also preserve the shape of the breast. However, a positive margin may result in an increased risk of local recurrences after BCT for any malignant tumour. Until now, the definition of a positive margin has been the subject of frequent debate (Hunt et al., 2014). In reality, surgeon removes the tumour is done by rough estimation of the boundary. Surgeon could not accurately determine the margin width until the pathologist makes a microscopic assessment. Pathologist's report might require a week or more to completed. If it shows the margins are not wide enough, the patient must undergo a second operation to remove the remaining malignant tissue. The operation would cause second physical and mental injury to patient.

For removing the tumour while minimizing the risk of leaving residual disease, many intra-operative methods have been proposed for tumour margin assessment, include optical coherence tomography (Nguyen et al., 2009), spectroscopy (Shipp et al., 2018) and molecular fluorescence imaging (Koller et al., 2018). In order to reduce the number of positive margins and offer surgeon real-time intra-operative information on the presence of positive resection margins, this study proposed a method which by using the specimen mammography during BCT.

Specimen mammography (McCormick et al., 2004) is routinely used to evaluate the surgical margin. The specimen is transported to a room near the operating room where mammography device was placed after the surgeon has resected the specimen. Mammogram is captured immediately and stored in a hospital diagnostic imaging system. The surgeon could quickly review the digital mammogram to assess the integrity of the resection. Specimen mammography with digital system have an edge on low equipment cost and less time consuming. The proposed method measured the pixel density and extracted the region of interest (ROI) by detecting the specimen boundary in mammography. Then the tumor boundary was detected by the proposed contouring methods from the ROI. A distance evaluation step was applied in a final stage to obtain the result.

## Methods

### Data acquisition

This study included 30 patients who received BCT. Each specimen mammogram is high resolution (over 1000 × 1000) and varies in size, has a corresponding ground truth image which is manually annotated by experienced surgeons. Two full field digital mammography (FFDM) systems were included in the study (GE Senographe Essential and Hologic Selenia Dimentions system). After wide excision of the tumor, location stiches were made on 12 (0º), 3(90º), 6 (180º) and 9 (270º) o'clock direction and clipped on the stiches in order to be easily identified on specimen mammogram. The pathologic report of margin distance was considered as the ground truth in this study.

### 2.3. Pixel density measurement, specimen boundary detection and ROI extraction

The proposed method measured the pixel density and extracted the region of interest (ROI) by detecting the specimen boundary in mammography. Then the tumor boundary was detected by the proposed contouring methods from the ROI. A distance evaluation step was applied in a final stage to obtain the result. Flow-chart of the proposed method is shown in Fig. 1. In the dataset, each tumor has a different proportion in the image. In order to accurately measure the distance between the tumor and the tissue, a standard one-dollar coin (20mm diameter) was placed in the specimen mammogram as a measuring scale. Pixel resolution was converted to millimeter by according the radius of the coin. (Figure 2)

Image pre-processing which removes the noise and enhances the quality of the image is very important step for image segmentation. The major problem with the precise segmentation of the specimen boundary is that the existence of noises which might affect the segmentation results. In order to suppress the noise in the background, an automatic thresholding method (Gonzalez and Woods, 2017) was performed to the specimen mammogram (Fig. 2.3(a)). In the proposed method, the connected component algorithm was utilized to extract the largest component (specimen), the form of labels, wedges, markers and some patient information in the background were discarded (Fig. 2.3(b)). In order to extract the tumour region completely, fill-hole operator was utilized to eliminate the black cavity. The morphological operators were used to smooth the boundary (Gonzalez and Woods, 2017), as shown in Fig. 2.3(c). The obtained specimen boundary was utilized as extract ROI for the following tumour boundary detection step (Fig. 2.3(d)).

**Margin width evaluation**

The distance between specimen boundary and tumour boundary was estimated as margin width. This study evaluated the margin width by the Euclidean distance (Deza, 2009). In an image coordinate plane, the distance between two points is usually given by the Euclidean distance (2-norm distance). The distance from a point to a line is the shortest distance from a fixed point to any point on a fixed line in Euclidean geometry.

**Tumour boundary detection**

The varying quality of specimen mammography makes tumor boundary detection becoming a difficult task. In order to overcome the conditions where specimen mammography has varied contrast, this study performed SegNet (Badrinarayanan et al., 2017) to contour the tumour in specimen mammography. SegNet is a deep convolutional network which address to image semantic segmentation for autonomous driving or intelligent robots. Due to the advantages of retaining high frequency details in the segmented images and also reducing the total number of trainable parameters in the decoders, SegNet has been used in medical image segmentation recently , such as gland segmentation in colon Cancer (Tang et al., 2018) and blood Cell Images Segmentation (Tran et al., 2018). The model is designed based on FCN. SegNet is composed of a symmetry network: the encoder and the decoder. The architecture of SegNet is shown in Fig. 3, it composed of three kinds of network: convolution, batch normalization and pooling. Convolution layers are used to extract local features; Batch normalization layers are used to expedite learning; and Pooling layers are utilized to down sampling feature map. Decoder aims to map the low-resolution feature maps from the encoder to obtain the same resolution as the input image feature map for pixel-level classification. The highlight of SegNet is that the decoder utilized max-pooling indices from the corresponding encoder stage to up-sample, this gives reasonably good performance and is space efficient

Data augment technique was used to create new images (Salehinejad et al., 2017) due to the smaller dataset,. In this work, 20 new images were generated from each case and resulting in 600 images. Combination of flipping, rotation, distortion and zoom transformations were performed randomly. SegNet utilized a pre-trained VGG16 model as the encoder part, thus could benefit from the features already created in the model and only focus on learning the specific decoding features. The proposed method used a mini-batch of 10 images, learning rate of 0.001 and the Adam optimizer. Finally, the morphological operator erosion was used to figure the obtained tumour boundaries.

**Platforms**

The traditional segmentation methods were implemented by Matlab (R2016a, MathWorks Inc., MA). The deep learning networks were trained on a Nvidia 1080Ti GPU. All methods were performed on a single CPU Intel i7 3.6 GHz personal computer with Microsoft Windows 10 operating system.

# Results

This study totally experimented 30 cases with manual sketched boundaries to evaluate the accuracy of the proposed method. In this work, measurement of pixel density was first performed to converted pixel resolution to millimetres and the specimen detection was applied to obtain the ROI. This study proposed the SegNet contouring approaches to obtain the tumour boundaries. Figure 5 demonstrates the final result applied the proposed contouring methods on a sample case.

Four practical similarity measures (Anbeek et al., 2004), i.e. the similarity index (SI), overlap value (OV), overlap fraction (OF) and extra fraction (EF) between the manually determined boundaries and the automatically detected boundaries, were calculated for quantitative analysis of the contouring results. The comparison of computational time consists of training time and testing time. The average difference between the proposed automatic segmentation and pathologist are listed in Table 1. Table 2 shows the four similarity measures of all cases. Average execution time of this approach is less than 5 seconds, which means the proposed system is suitable for intra-operative tumor margin evaluation. In this study, the safety margin width recommended as 10mm. When the margin width is less than 10mm, the system would mark the area with a caution. Figure 6 illustrates the estimation and evaluation results.

## Discussion

Although the plot of testing accuracy always close to training accuracy, the plot of training loss continues to decrease with experience and the validation loss decreases to a point and begins unstable. This situation could be identified as an overfitting model. There were many reasons give rise to overfitting, we inferred that is due to the size of dataset is small. Thus, this study expected the specimen mammography dataset would be expanded much more, then the algorithm has the potential to come up with a better model in the future.

## Conclusions

During the surgery, the doctor usually measures the margin width to ensure that the tumour is removed clearly. A number of assisted diagnostic systems for measuring tumour margins have been proposed, however these methods require high cost equipment or contrast medium injection for patient. This study proposes a fast, low-cost computer-aided method for detecting tumour boundaries and estimating margin width. The experimental results revealed that the average difference of deep-learning techniques is similar to doctor's manual sketching, which means the deep-learning techniques could sketch boundary reasonably. That is the deep-learning techniques have the opportunity to automatically find new features without human intervention. With the aid of deep learning techniques, the proposed scheme would be a potential procedure in intra-operative measurement system

## Authors' contributions

Author contributions: Y.-L. Huang, H.-K. Wu and D.-R. Chen designed research; W.-C. Shia conducted review and editing; Y.-L. Huang and D.-R. Chen provided funding acquisition, project administration, and resources; and Y.C. Chen, W.-C. Shia and Y.-L. Huang wrote the paper.

## Acknowledgements

This work was supported by the Ministry of Science and Technology, Taiwan, Republic of China, under Grant MOST 109-2221-E-029-024.

# Figures

**Figure 1 - Flow-chart of the proposed method**

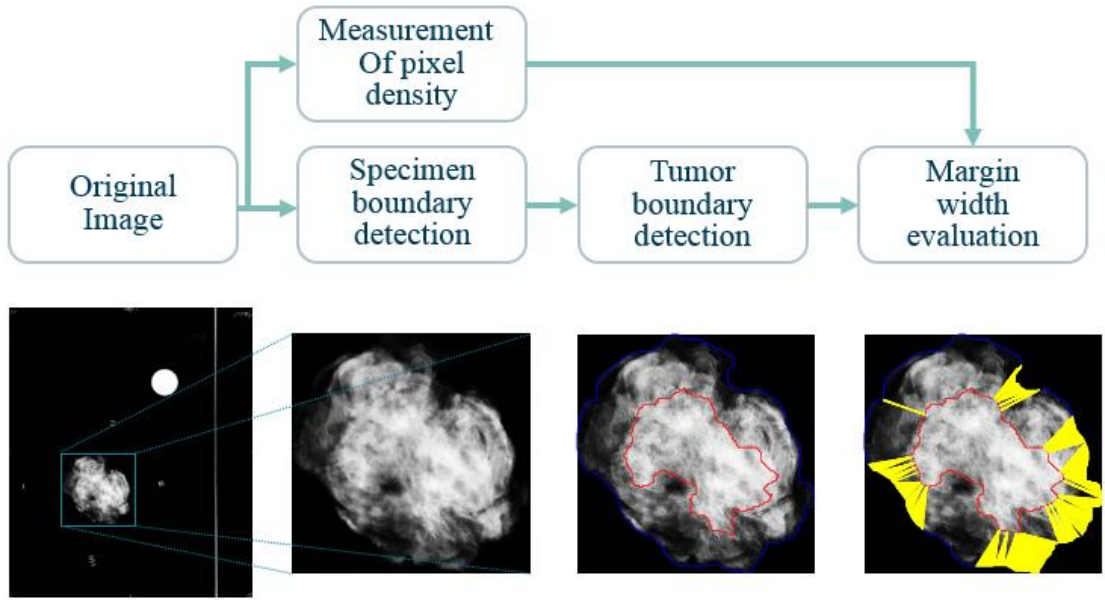

**Figure 2 - Measurement of pixel density (a) The case with large pixel density (239 pixels per 1mm) and (b) the case with small pixel density (104 pixels per 1mm)**

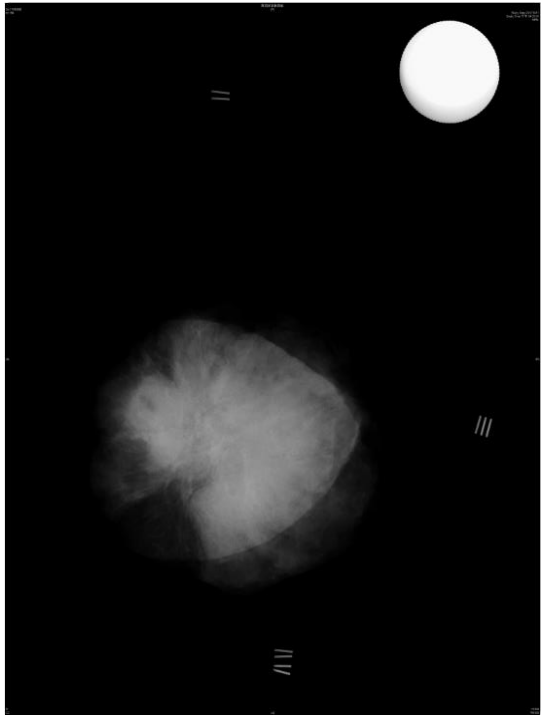
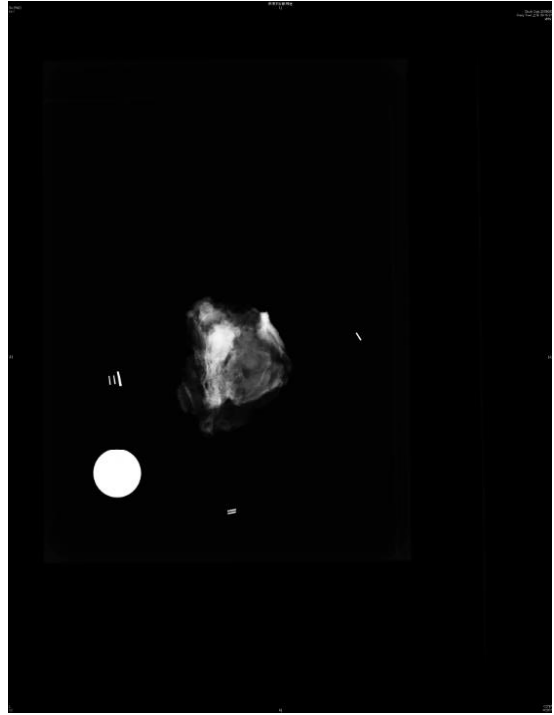

(a)          (b)

**Figure 3 - Result of specimen boundary detection and ROI extraction (a) binary image, (b) specimen region, (c) extracted specimen boundary (red) and (d) extracted ROI**

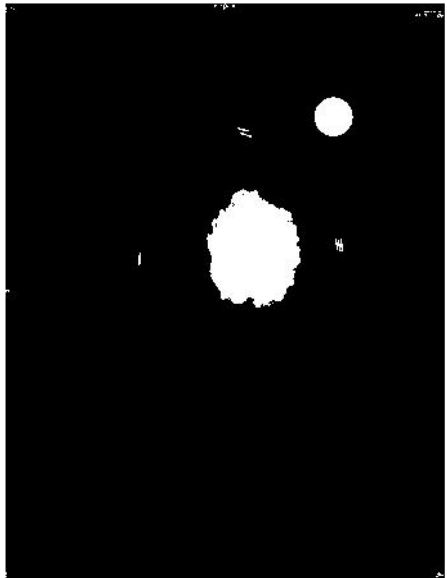
(a)

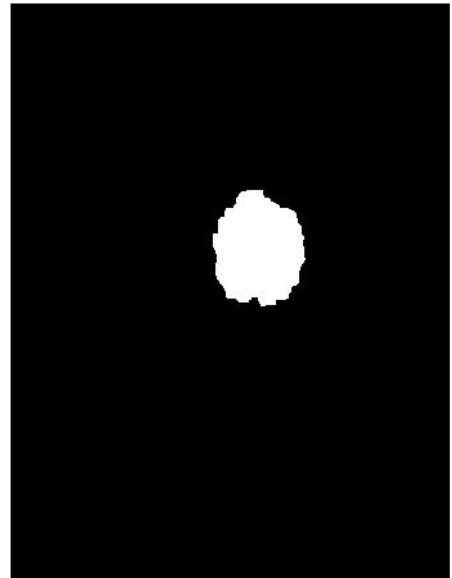
(b)

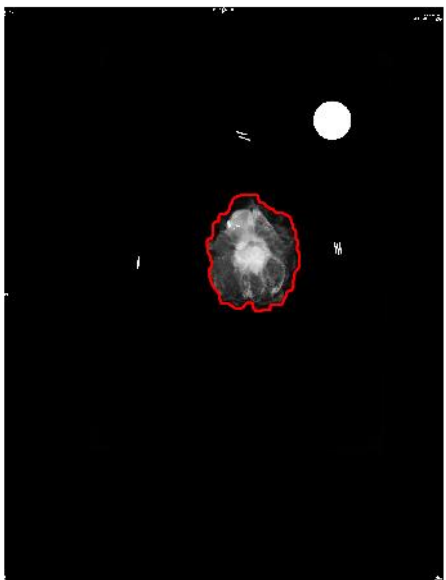
(c)

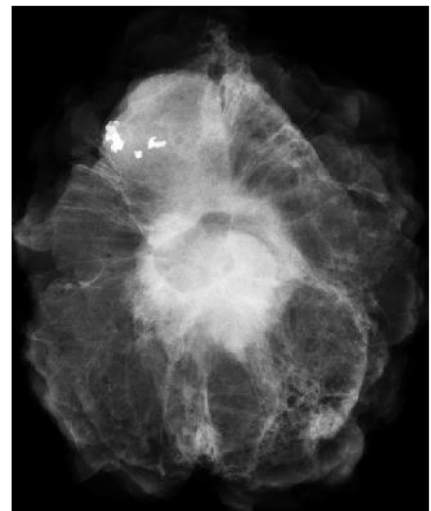
(d)

**Figure 4. SegNet architecture**

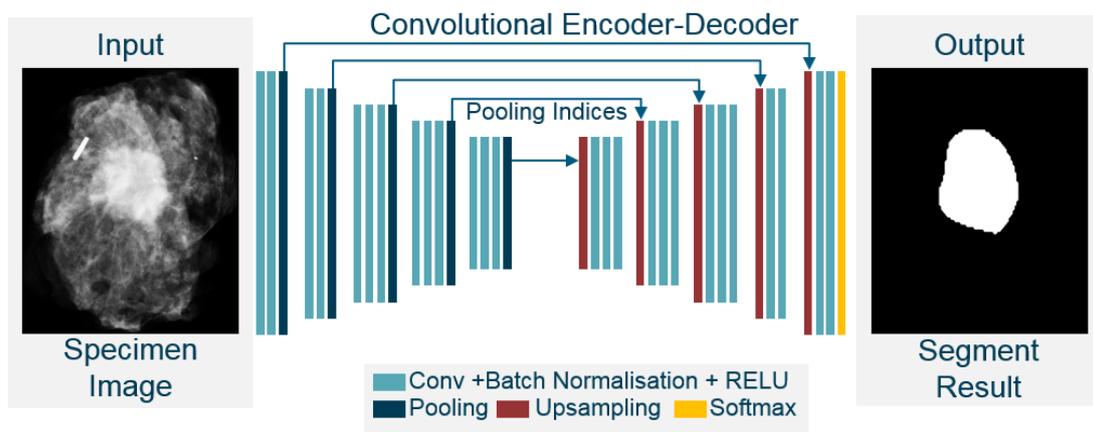

**Figure 5. Presentation Case (a) original image, (b) extracted ROI and specimen boundary (yellow), (c) final segmentation result and (d) evaluated margin width between the proposed contouring method and the manual sketching by physician.**

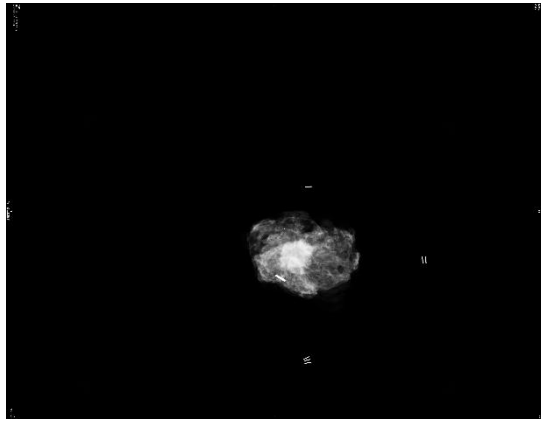
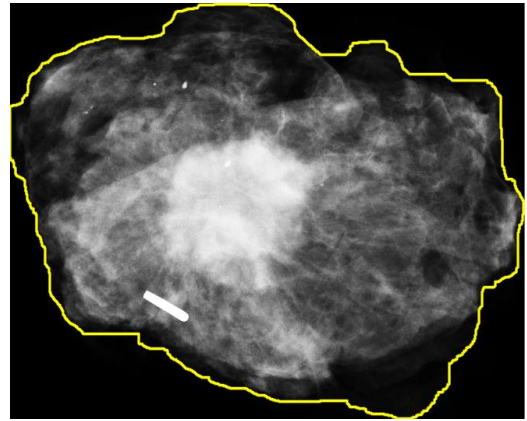

(a)                  (b)

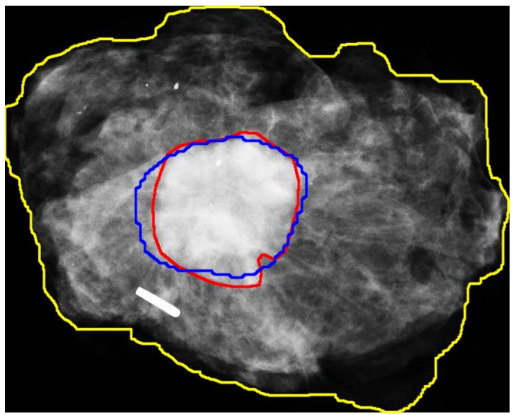
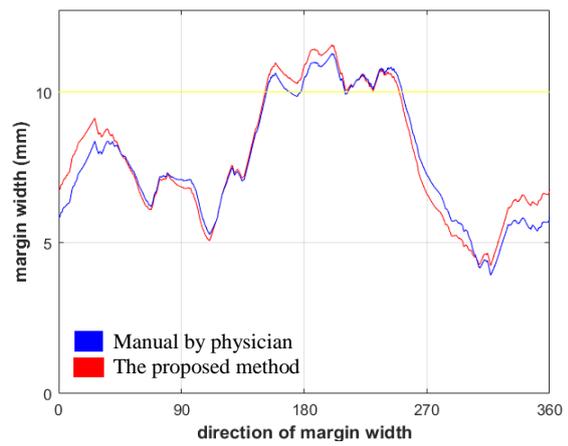

(c)                  (d)

**Figure 6**. Evaluation results (a) extracted specimen boundary (blue), tumor boundary(red), the region less than the predefined safety margin width (yellow) and (b) the evaluated margin width

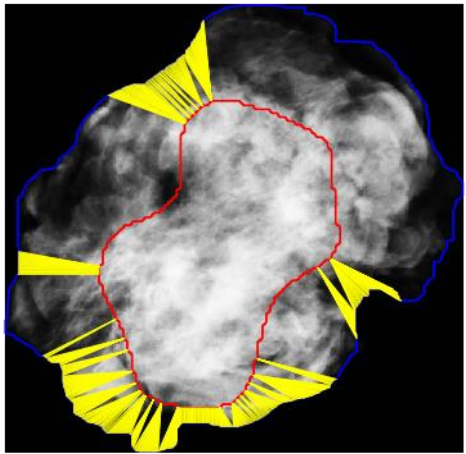
(a)

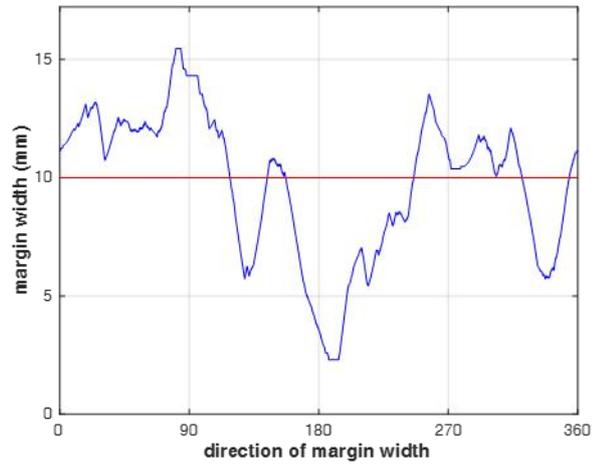
(b)

# Tables

**Table 1 - The results of proposed method compared with pathology margin width**

| Pathology direction | Manual sketch | SegNet |
|---|---|---|
| 3 o'clock | 7.31± 6.25 | 7.29± 6.32 |
| 6 o'clock | 5.44± 4.11 | 5.09± 4.76 |
| 9 o'clock | 6.96± 5.85 | 6.18± 5.48 |
| 12 o'clock | 6.78± 6.68 | 7.56± 6.81 |
| Average | **6.62± 5.72** | **6.53± 5.84** |

**Table 2 - The similarity results of the proposed method which compared with the physician manual sketching**

| | SegNet | | | |
|---|---|---|---|---|
| Case # | SI | OF | OV | EF |
| 1 | 0.91 | 0.94 | 0.84 | 0.12 |
| 2 | 0.72 | 0.96 | 0.56 | 0.70 |
| 3 | 0.65 | 1.00 | 0.48 | 1.07 |
| 4 | 0.80 | 0.96 | 0.67 | 0.44 |
| 5 | 0.90 | 0.93 | 0.82 | 0.13 |
| 6 | 0.74 | 0.83 | 0.59 | 0.41 |
| 7 | 0.55 | 0.38 | 0.38 | 0.00 |
| 8 | 0.73 | 0.57 | 0.57 | 0.00 |
| 9 | 0.86 | 0.95 | 0.76 | 0.25 |
| 10 | 0.75 | 0.61 | 0.60 | 0.01 |
| 11 | 0.54 | 1.00 | 0.37 | 1.72 |
| 12 | 0.93 | 0.97 | 0.87 | 0.12 |
| 13 | 0.85 | 0.91 | 0.73 | 0.24 |
| 14 | 0.42 | 0.97 | 0.26 | 2.66 |
| 15 | 0.39 | 1.00 | 0.24 | 3.14 |
| 16 | 0.25 | 1.00 | 0.14 | 5.92 |
| 17 | 0.75 | 1.00 | 0.60 | 0.66 |
| 18 | 0.78 | 0.98 | 0.63 | 0.55 |
| 19 | 0.72 | 0.70 | 0.57 | 0.24 |
| 20 | 0.43 | 0.27 | 0.27 | 0 |
| 21 | 0.69 | 0.98 | 0.52 | 0.87 |
| 22 | 0.36 | 0.22 | 0.22 | 0.00 |
| 23 | 0.80 | 0.98 | 0.66 | 0.48 |
| 24 | 0.84 | 0.74 | 0.73 | 0.02 |
| 25 | 0.84 | 0.75 | 0.73 | 0.04 |
| 26 | 0.87 | 0.79 | 0.77 | 0.03 |
| 27 | 0.38 | 1.00 | 0.23 | 3.27 |
| 28 | NaN | NaN | NaN | NaN |
| 29 | 0.28 | 0.93 | 0.16 | 4.65 |
| 30 | 0.61 | 0.92 | 0.44 | 1.09 |
| **Average** | **0.62** | **0.73** | **0.49** | **0.73** |